# Quantum aspects of high dimensional formal representation of conceptual spaces

M S Ishwarya, Aswani Kumar Cherukuri[*](cherukuri@acm.org)
School of Information Technology, Vellore Institute of Technology, India.

**Abstract:** Human cognition is a complex process facilitated by the intricate architecture of human brain. However, human cognition is often reduced to quantum theory based events in principle because of their correlative conjectures for the purpose of analysis for reciprocal understanding. In this paper, we begin our analysis of human cognition via formal methods and proceed towards quantum theories. Human cognition often violate classic probabilities on which formal representation of conceptual spaces are built. Further, geometric representation of conceptual spaces proposed by Gärdenfors discusses the underlying content but lacks a systematic approach (Gärdenfors, 2000; Kitto, Bruza, & Gabora, 2012). However, the aforementioned views are not contradictory but different perspective with a gap towards sufficient understanding of human cognitive process. A comprehensive and systematic approach to model a relatively complex scenario can be addressed by vector space approach of conceptual spaces as discussed in literature. In this research, we have proposed an approach that uses both formal representation and Gärdenfors geometric approach. The proposed model of high dimensional formal representation of conceptual space is mathematically analysed and inferred to exhibit quantum aspects. Also, the proposed model achieves cognition, in particular, consciousness. We have demonstrated this process of achieving consciousness with a constructive learning scenario. We have also proposed an algorithm for conceptual scaling of a real world scenario under different quality dimensions to obtain a conceptual scale.

## 1. Introduction:

Humans create memory in accordance to their survival relevance in the environment (Gunji, Sonoda, & Basios, 2016). These memories are often associated as concepts based on neuro-computational correlates happening at high level cognitive functions of human brain (Reggia, Katz, & Huang, 2016). The information represented in concepts along with the conceptual space continuously evolve over time during cognitive processes (Li, Zhang, Song, & Hou, 2016). These cognitive processes are defined as maps between the instance (cues) and the representations of instance (Gunji et al., 2016). The mental maps represent association of information distributed across the dimensions of conceptual space (Bolt et al., 2016). Cueing the mental map implies the process of understanding (measurement) of concepts at a given time granule since mental maps tend to evolve the measurement at differs at different time granule. Different lobes of human brain handle different kinds of information. For example, occipital lobe of human brain handles visual cues while parietal lobe handles taste, touch and temperature. Information is distributed across human brain and mapped via mental maps based on their recruitment of neuronal gaps (Gunji et al., 2016) on presenting a particular cue. Every cue is associated with a corresponding description and recorded as a concept. The mental maps that are formed interconnected concepts are called conceptual space (Tversky, 1993).

When a conceptual space is presented with successive plural stimulus with regard to a scenario leading to non-algorithmic jumps, the ensemble shows quantum-like characteristics such as superposition, entanglement, etc. (Gunji et al., 2016). Successive pieces of cues with regard to scenario, the composition and decomposition of memory window explains quantum conjectures(F T Arecchi, 2013). A judgement in a conceptual space cannot be performed using classical Bayes and Inverse Bayes inferences in reality (F Tito Arecchi, 2011). Bayes and Inverse Bayes inferences restrict the human cognitive judgements to classical probabilities. Conceptual spaces require a more generalized probability rather than a classical probability theory such as quantum probability. Judgements in a conceptual space modifies its states with a novel, enriched and more appropriate version of the previous instance of the state (F T Arecchi, 2013). This explains the fact that the information represented in the concepts along with the conceptual space continuously evolves over time. Introspection should collapse all possible versions of existence of a concept to a definite state in conceptual space (Farrell & McClelland, 2017). A subset of the representation of a stimulus introduces a view. Judgement varies accordingly when considering different views. Throughout this paper, we may switch between the terms 'conceptual space' with 'mental ensemble' and 'concept' with 'mental state' with regard to the context in which it is used. We use 'conceptual space' and 'concepts' in context of formal representation and 'mental states' and 'mental ensemble' while adapting the formal representation in a biological context. However, these alternate terms means the same but in different context.

Quantum theories need not necessarily model microscopic situations; they might as well used for modelling classical formal representation(Aerts, Gabora, & Sozzo, 2013). The attempt for modelling classical macroscopic situations until recently has reported the evidences that quantum theories can appropriately model the process of human cognition. When quantum theories are applied for human cognitive models the concepts are regarded as the '*entity in specific state*' rather than '*container of instances*'(Aerts et al., 2013). Hence, we regard the concepts as mental states and the conceptual space formed by the concepts as mental ensemble. In quantum theory, the states are unit vectors and their transitions are represented unitary matrices using quantum gates. However, this kind of representation and transitions can happen only in ideal cases since it does not consider the real world processes such as cues from the environment and irreversible transitions with regard to a cognitive system (Sergioli, Santucci, Didaci, Miszczak, & Giuntini, 2018). It is therefore identified that quantum theories may well be used as a treatment for human cognitive processes with little adaptation of modified notions in both the theory.

Among the cognitive processes that exist in human brain, our particular interest is on consciousness considering the strong correlation between consciousness and quantum theories as suggested by literature (Baars & Edelman, 2012; Beck & Eccles, 1992; Fields, Hoffman, Prakash, & Singh, 2018). According to Penrose and Hameroff (2014) consciousness in human brain is a result of quantum processes happening in collections of microtubules with neurons that regulate the synaptic activity. The microtubules are interconnected at dendrites and collapse of superposition states in microtubules produces spikes at axon. Quantum states evolve to conscious state due to transformations caused by the application of quantum gates. Beck and Eccles (1992) proposed that cognition occurs at dendrites and quantum physics is involved in chemical process called exocytosis that causes polarization of synaptic connections of a neuron. These two theories with regard to consciousness and quantum theories regard as quantum brain models of consciousness (Bruza & Busemeyer, 2012). One important aspect of consciousness is that it happens during the process of experiencing the cue in a mental ensemble based cue introduced to it. Consciousness is temporal phenomenon and it facilitates learning by deciding between the conscious (definite) and uncertain states (Baars & Edelman, 2012).

Consciousness is often defined as '*difference that makes a difference*' with regard to neuronal interaction that happens at a massive level. When a measurement about the knowledge is made to a conceptual space, we observe the states that exist just before measurement since measurement process itself causes a difference. A concept space is measured by cue or an interaction with the environment. All neuronal processes need not become conscious (Baars & Edelman, 2012). Consciousness is special kind of brain event. Consciousness is a fundamental problem of quantum physics (S. Hameroff & Penrose, 2014). However, the current quantum level proposal of consciousness does not explicitly and sufficiently explain the process of consciousness. Consciousness can be viewed as the inner knowledge of the subject. Consciousness at conceptual level is highly required considering the process of consciousness is subjective (Gök & Sayan, 2012). Consciousness occurs in two forms based on the information being processed namely, primary and higher order consciousness. Primary consciousness process sensory data while the higher order consciousness process symbolic information learnt earlier. Higher order consciousness deals with the processing of information with symbolic, abstract and language dependent data. Our particular interest lies in modelling higher order consciousness (Baars & Edelman, 2012). Consciousness is biased over a more informative cue while it fades for a weak or redundant cue. Consciousness is attributed to self since it observes itself (knowledge in a mental ensemble). The study of consciousness should indeed explain about conscious and uncertain mental states. During the process of consciousness in the neuronal brain structures, quantum phenomenon is exhibited (Gök & Sayan, 2012).

It can be observed that conceptual spaces, quantum theories and consciousness own reciprocal relations with each other based on the aforementioned common conjectures. Considering these correlations, we propose the following objectives in this research,

1. High dimensional representation of conceptual spaces that combines both formal and geometrical representation of conceptual spaces exhibit quantum aspects.

2. Cognition, in particular higher order consciousness is achieved by this proposed high dimensional formal representation of conceptual spaces.

3. An algorithm for conceptual scaling of a real world scenario to obtain a formal context is proposed

Rest of the paper is organized as follows. In the next section of the paper, we provide brief descriptions of quantum principles and processes in terms of definitions followed by the description on conceptual spaces. In section 3 of the paper, we derive high dimensional mental ensemble that exhibits quantum aspects. In section 4, we proposed a formal method to achieve consciousness in aforementioned high dimensional mental ensemble. In section 5, we adapt this derived high dimensional mental ensemble for a constructive scenario to exhibit cognition. Later sections of the paper provide discussions and conclusions of the proposed work.

**2. Background:**

Considering the reciprocal relationship between the quantum theories and conceptual spaces as mentioned in the previous section, we elaborate the fundamentals of each in this section of the paper. We first brief the fundamental of quantum theory followed the fundamentals of conceptual spaces with respect to formal representation and Gärdenfors representation based on which we present our proposed formal method of high dimensional conceptual space that achieves cognition.

**2.1 Fundamentals of Quantum Theory:**

A system that follows quantum principles is regarded as quantum system (Gruska & Republik, 2004). Information in a quantum system is stored in qubits according to quantum theory in which a qubit can exist in of $|0\rangle$ and $|1\rangle$ simultaneously where $|0\rangle$ and $|1\rangle$ are the basis states of single qubit system. This principle of quantum theory that allows a qubit to exist simultaneously in 0 and 1 is called superposition (E.Knill, R.Laflamme, & H.Barnum, 2011; Nielsen & Chuang, 2010). A quantum state is represented by a vector $|\psi_{(t)}\rangle$ described the linear superposition of basis states. Let the set *B = {i / i denotes a state}* is a set of basis states, if for all $i, j \in B \langle i, j \rangle = \delta_{ij}$, that is

$$\langle i | j \rangle = \begin{cases} 1; i = j \\ 0, otherwise \end{cases} \quad (1)$$

and for any initial state X and final state Y it holds

$$\langle Y | X \rangle = \sum_{i \in B} \langle Y | i \rangle \langle i | X \rangle \quad (2)$$

Equation 1 states that basis states can take values 0 or 1 and the equation 2 states that a final state *Y* can be formed from the initial state *X* only if they share a common basis vectors. The amplitude of an event is a sum of amplitudes of an event corresponding to the given set of basis states. For any set B of basis states and for any initial state X

$$\sum_{i \in B} |\langle i | X \rangle|^2 = 1 \quad (3)$$

Hilbert space is mathematical framework that is best suitable for describing concepts, principles, laws and processes with regard to quantum mechanics while state can be represented as vectors in Hilbert space (Gruska & Republik, 2004). Let H be a Hilbert space of states of a quantum system. An Observable $O = \{E_1, E_2, ...... E_k\}$ is a set of disjoint and mutually orthogonal subspaces such that

$$H = E_1 \oplus E_2 \oplus ....... \oplus E_k \text{ (an orthogonal sum)} \quad (4)$$

In order to observe quantum information from quantum systems, measurement has to be carried out. Measurement destroys the superposition of the qubit to one of basis states. Let $|\phi\rangle$ be the state and $O = \{E_1, E_2, ...... E_k\}$ be the observable. $|\phi\rangle$ can be expressed uniquely as a linear superposition of components along with each $E_i$

$$|\phi\rangle = \sum_{i=1}^{k} \alpha_i | \phi E_i \quad (5)$$

Another important principle of quantum theory is entanglement. Two particles are said to be entangled when quantum state of one of the particle cannot independently described without the quantum state of the other particle even when they are placed farther apart. Orthogonal states can only be entangled. For example, a $\langle 0|$ state can be entangled with $|1\rangle$ state. Alternatively, $|1\rangle$ state can be entangled with $\langle 0|$.

Having introduced the fundamentals of the quantum theory, we now proceed with the basics of the conceptual spaces in the next section of the paper considering the correlations between the human cognitive processes and the quantum theories as suggested by the literature (Aerts, 2007; Aroyo & Amsterdam, 2017; Baars & Edelman, 2012; Gunji et al., 2016; Kitto et al., 2012; Masuyama, Loo, & Kubota, 2014). We mathematically deduce a high dimensional formal representation of conceptual spaces from formal representation and Gärdenfor's geometrical framework of conceptual spaces and analyse its correlation with quantum system.

**2.2. Fundamentals of Conceptual spaces:**

Human cognition process is conventionally described and represented in the form of conceptual space (Kumar et al., 2015). Concepts are basic entities of conceptual spaces while the relation between such concepts accounts for cognitive processes (Gärdenfors, 2000). A concept C:=(O, A) is a pair of object and the attributes that describes the objects. Each object can be real world entity, instance, event, etc. depending on the context of the conceptual space. A conceptual space $C$ is spanned by a set of quality dimensions $D$ that best describes the concepts in $C$. Each quality dimension $d_i \in D$ is a meaningful representation of concept $c_i \in C$ in terms of set of attributes such that concept in $C$ can be judged for similarity based on dimensions in $D$. Every concept that is learnt from human cognition is either categorized based on a difference or grouped based on similarity. Formal representation and Gärdenfors geometric representation of conceptual spaces are two different views of interest on conceptual spaces. According to Gärdenfors, a conceptual space possess following criterion,

**Criterion P:** A natural concept is a convex region of domain in a conceptual space

**Criterion C:** A natural concept is represented as a set of convex regions in a number of domains together with an assignment of salience weight to the domains and information about how the regions in different domains are correlated.

Criterion P and C mentioned above explains the nature of concepts in a concept lattice. According to formal representation of conceptual spaces, formal context $K = (G, M, I)$ is called the conceptual scale with regard to a real world scenario R (Belohlavek, 2008). Formal context represents the scenario in the term of data table which conventionally holds objects in rows and attributes in columns while the entries in the table represents the binary relation between the objects and attributes. A concept $C$ in a formal context $K$ is represented by a pair $(X, A)$ where $X \subseteq G, A \subseteq M$ such that,

$$x^{\uparrow} = \{a \mid (x,a) \in I, \forall x \in X, a \in A\}$$
$$a^{\downarrow} = \{x \mid (x,a) \in I, \forall x \in X, a \in A\} \qquad (6)$$

The aforementioned explanation with regard to concepts and conceptual spaces are static and they do not record the transitions that are happening in the conceptual space. Conceptual Time System(Woff, 2011) are a variant of formal methods of conceptual analysis that allows the notion of distributed objects that has traces of concepts at certain time granules. The intention behind this import of Conceptual Time System represent the state of the object with a formal representation at a given time granule as well the transition of states with respect to the object of interest. Conceptual Time System differs from other formal methods because of adding a temporal phenomenon to the conceptual space to represent the dynamically changing concepts and their links in conceptual space.

User cognitive state and information continuously evolve over time. Despite advancements in information retrieval, capturing and understanding the dynamic information trajectories is still challenging. Quantum Language Model exploits associations to model the trajectories in human cognitive state and information (Li et al., 2016). Human cognition has two important processes such as apprehension and judgement. The former deals with the recruitment of neuronal gaps while later deals with recalling the most associative gap in a way similar to superposition and measurement(F T Arecchi, 2013). Adding to these, fuzzy inferences of bidirectional associative memory satisfies quantum postulates (Masuyama et al., 2014). The functionalities of bidirectional associative memory such as learning, memorising and recalling the cue and its associated pattern can well be modelled using formal representation of conceptual spaces (Aswani Kumar, Ishwarya, & Loo, 2015). According to (Kitto et al., 2012), sematic space models are too situation dependent and relevant to text collection from which it is constructed. Hence, text and corpus based information is transformed to concept and property inspired quantum approach. Quantum inspired approach not only justifies and reciprocates human cognition but machine learning techniques as well. Another literature on quantum based approach for machine learning problem asserts that quantum based approach can be used for macroscopic situation as well (Sergioli et al., 2018) and need not necessarily model microscopic situations. Adding to these, quantum based cognition is employed to achieve high level reasoning (Bruza et. al, 2018), decision making (BroeKaert et. al, 2018), selecting a strategy (Busemeyer, 2018) and adaptive estimation (Busemeyer, 2018)

Human cognitive processes such as perception and consciousness select their habitable space-time via quantum measurements (Igamberdiev & Shklovskiy-Kordi, 2017). According to Busemeyer and Wang, quantum processes are of two types namely physical and mathematical. However, human cognitive are non-physical mathematical quantum processes (Busemeyer & Wang, 2014). Aerts make it further evident that concepts can be modelled using mathematical structure of quantum mechanics by representing concepts in complex Hilbert space (Aerts, 2007). Adding to the strong correlation between quantum mechanics and human cognition(Aerts, 2009; Bolt et al., 2016; Busemeyer, Fakhari, & Kvam, 2017; Melkikh & Khrennikov, 2015; Reggia et al., 2016), literature suggests a direct correlation between consciousness a special case of cognition and quantum mechanics(Farrell & McClelland, 2017; O'Rourke, 1993). Consciousness is a special brain event and its origin is still ambiguous due to its subjective nature (Beck & Eccles, 1992). Quantum models are prepared and measured with a view on better representation of conceptual spaces (Kitto et.al, 2018).

According to (S. R. Hameroff, Craddock, & Tuszynski, 2014; Overgaard & Mogensen, 2017) consciousness is regarded as stable knowledge with regard to a scenario at a given point of time. As mentioned elaborately in the previous section of the paper, consciousness is strongly correlated with quantum mechanics (Gök & Sayan, 2012; S. Hameroff & Penrose, 2014; Mensky, n.d.; Reggia et al., 2016).

Considering the strong reciprocal correlations between quantum mechanics, cognition and correlation, we propose model the mathematical quantum process in conceptual spaces for the following identified research gap. Gärdenfors representation of conceptual spaces can explain the dynamics of conceptual spaces better with a systematic approach (Kitto et al., 2012). Similarly, the formal representation of conceptual spaces although has a formal approach in treating conceptual spaces, it is confined to classic probability (Belohlavek, 2008). Human cognitive processes violate classic probabilities (Bruza & Busemeyer, 2012). In this paper, we derive an approach for the treatment of conceptual spaces by the combined approaches of formal representation and geometrical representation of conceptual spaces. By doing this combination, conceptual spaces are represented in vector space models. It is observed during our research that such high dimensional vector space models of conceptual spaces exhibits quantum aspects. In the further section of the paper, we have achieved cognition, in particular consciousness via proposed high dimensional formal representation of conceptual spaces. In this process of achieving of consciousness, it is observed that the model continued to exhibit quantum like characters. In the next section of the paper, we have derived a high dimensional formal representation of conceptual spaces.

**3. High dimensional mental ensemble with quantum aspects:**

Formal representation of conceptual spaces and Gärdenfors framework of conceptual spaces are connectionist level and conceptual level representation of cognitive information respectively. These representations are just different views on cognitive information rather than being contradictory. Gärdefors approach of conceptual spaces can explain conceptual spaces better with a systematic approach while the implementation for complex examples are not clear (Bruza & Busemeyer, 2012). A systematic approach for representing complex scenario is required. In this section of the paper, we build a high dimensional representation of conceptual space (mental ensemble) with the help of both formal representation and Gärdenfors representation of conceptual spaces. Further, we represent the conceptual space representation to vector based conceptual space for the systematic representation of complex scenario in conceptual space.

Let a real world scenario *R* is represented based on the instances (extent) *M* and its description (intent) *G*. Conceptual scaling is performed on the real world scenario *R* to convert it to a context represented by a triple *K=(M,G,I)* where *I* represents the relation between event *M* and description *G*. A formal concept *C* with regard to scenario *R* is a pair *(X, A)* where $X \subseteq M$ and $A \subseteq G$ can be described as in equation 6.

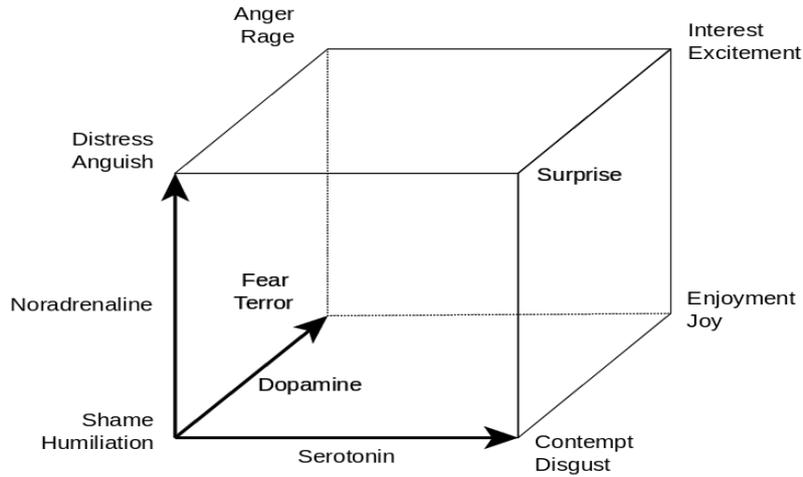

Figure 1: Lövheim cube of emotion projecting emotions in 3-D space (Lövheim, 2012)

According to Gärdenfors, a concept is described by set of related attributes of various quality dimensions (Gärdenfors, 2000). Adapting this definition of concepts, we rewrite the equation 6 in terms of quality dimensions and attributes. Let *D* be the set of all quality dimensions such that $D = \sum_{i=1}^{n} d_i$ considered in the real world scenario *R* where $d_i$ is i$^{th}$ quality dimension in set D.

$$x^{\uparrow} = \sum_{i=1}^{n} \sum_{j=1}^{m} d_i a_j I_j \qquad (7)$$

Equation 7 represents an instance $x^{\uparrow}$ by linear combination of attributes from their respective quality dimensions that are related to it with respect to binary relation *I* in formal context *K*. In equation 7, *n* represents the number of quality dimensions and *m* represents the number of attributes present in particular quality dimension $d_i$. For example, Fig. 1 represents a three dimensional emotional conceptual space projected with regard to quality dimensions such Dopamine (polar dimension), Serotonin (linear dimension) and Noradrenaline (linear dimension) as in equation 7. Each dimension is a geometrical structure. Any emotion that falls into this conceptual space can be represented in terms of the quality dimensions such as Dopamine, Serotonin and Noradrenaline. The projection of each quality dimension varies according to an emotion. In other words, each emotion can be represented by the collection of projection from individual quality dimensions. For example, in Fig. 1, quality dimensions 'Noradrenaline' and 'Dopamine' contribute more than 'Serotonin' for the emotion anger. Each hormone may have N number of attributes related to it depending on the scenario. For example, hormone 'Noradrenaline' may have attributes like heart rate, blood pressure, energy release, blood flow, gastrointestinal mobility etc. since 'Noradrenaline' effects on body causes increases in heart rate, increase in blood pressure, increase in blood flow, decrease in gastrointestinal mobility and release of glucose from energy stores.

$$x^{\uparrow} = \sum_{i=1}^{n} d_i \begin{pmatrix} a_1 I_1 \\ a_2 I_2 \\ . \\ . \\ a_m I_m \end{pmatrix} \quad (8)$$

Similarly, in equation (9), $a_n I_n$ represents a projection in a subspace within one quality dimension $d_i$. $x^{\uparrow}$ is the linear combination of projections in multi-dimensional subspaces. We further expand equation 8 for better understanding as shown in equation 9

$$A = d_1 \begin{pmatrix} 1 \\ 0 \\ 0 \\ . \\ . \\ 0 \end{pmatrix} + d_2 \begin{pmatrix} 0 \\ 1 \\ 0 \\ . \\ . \\ 0 \end{pmatrix} + .... + d_n \begin{pmatrix} 0 \\ 0 \\ 0 \\ . \\ . \\ 1 \end{pmatrix} \quad (9)$$

Since we are considering a finite set of objects and attributes in a context *K*, the projections on the subspaces with regard to each quality dimension are finite as well. If $x^\uparrow$ owns an attribute (projection in a particular subspace) the term $a_n I_n$ is 1 or if $x^\uparrow$ does not own an attribute (projection in a particular subspace) the term $a_n I_n$ is 0.

Let *S* be a quantum system and $|\psi\rangle$ be a quantum state in a quantum system *S*. The quantum state $|\psi\rangle$ is obtained from the set of basis vectors $\{|0\rangle, |1\rangle\}$. The basis vectors are orthogonal and they represents the possible states of the qubit. The possibility that a qubit might settle in any one of the state in basis vector states is represented by probability amplitude of that qubit. For example, equation 10 represents a quantum state $\psi$ of quantum system with a single qubit.

$$\psi = \alpha \begin{bmatrix} 1 \\ 0 \end{bmatrix} + \beta \begin{bmatrix} 0 \\ 1 \end{bmatrix} \qquad (10)$$

In equation 10, $\alpha$ and $\beta$ are the probability amplitude that the state $|\psi\rangle$ will settle at the one of the basis vectors. $\alpha$ represents the probability that $|\psi\rangle$ will settle at $|0\rangle$ and $\beta$ represents the probability that $|\psi\rangle$ will settle at $|1\rangle$ such that $|\alpha|^2 + |\beta|^2 = 1$. It can be observed that $|\alpha|^2 + |\beta|^2 = 1$ is unit sphere where $\alpha$ and $\beta$ are Hopf coordinates formed based on the angle of projections $\theta$ and $\varphi$ as shown in Fig. 2 where $\alpha = \cos\frac{\theta}{2}$ and $\beta = e^{i\phi}\sin\frac{\theta}{2}$. On taking a closer look of Fig. 2, $\alpha$ and $\beta$ are the degrees of freedom that rotates the basis states in Bloch sphere. A single qubit has two degree of freedom, while two qubits can have four degrees of freedom and so on. We represent the qubit in ket notation to proceed with the derivation. For example, $|0\rangle$ is the ket representation of column vector $\begin{bmatrix}1\\0\end{bmatrix}$ and $|1\rangle$ is the ket representation of the column vector $\begin{bmatrix}0\\1\end{bmatrix}$. The quantum state $|\psi\rangle$ of the quantum system $S$ can be represented by $2^N$ basis state vector where $N$ is the number of qubits in the system. The quantum state $|\psi\rangle$ of the quantum system $S$ with $N$ number of qubits represented in ket notation is shown in equation 11.

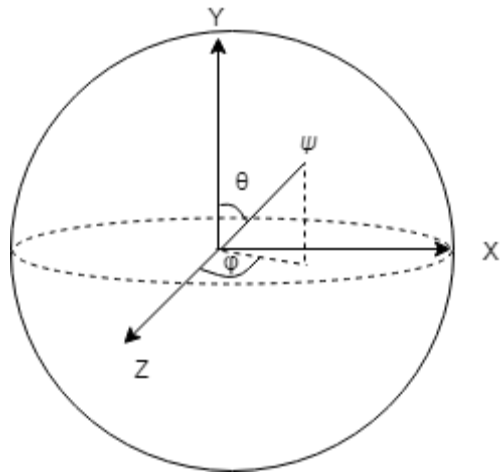

Figure 2: Degrees of freedom and state representation with regard to a qubit (Bloch Sphere)

$$|\psi\rangle = \alpha\begin{bmatrix}1\\0\\.\\.\\0\end{bmatrix} + \beta\begin{bmatrix}0\\1\\.\\.\\0\end{bmatrix} + ...... + \delta\begin{bmatrix}0\\0\\.\\.\\1\end{bmatrix} \quad (11)$$

It is interesting to note that an object representation of concept in a high dimensional mental conceptual space represented in equation 9 is analogous to quantum state representation in $N$ qubit quantum system $S$ in equation 11. It is further interesting to note that the degree of freedom is the dimension that basis vector can be projected. This is analogous as well with attributes of the concept taking 0 or 1 values based on the projection with regard to the quality dimension. An object in a concept is a linear combination of attributes from their respective quality dimensions where as a quantum state is a linear superposition of basis vectors from their respective degrees of freedom as shown in Fig. 1 and 2. The state in mental ensemble as well as quantum system varies based on the contribution of projections from individual quality dimensions and displacement of basis vector with regard to the degrees of freedom respectively as conveyed by Fig. 1 and 2.

From this derivation of object representation (mental state) high dimensional conceptual space (mental ensemble) analogous to quantum states represented by $N$ qubits, following notions in respective context are inferred equivalent,

1. An attribute in a high dimensional mental ensemble is equivalent to basis vectors in quantum system.

2. A quality dimension in a high dimensional ensemble is equivalent to degree of freedom of the basis vector.

3. A mental state in a high dimensional mental ensemble is equivalent to quantum state in quantum system. A mental state is represented by the linear combination of the attribute in different quality dimension while a quantum state is represented by linear superposition of basis vectors with different degrees of freedom.

On adapting the aforementioned inferences to Fig. 1, the attributes under each quality dimension (for example, the attribute of 'Noradrenaline' are heart rate, blood pressure, blood flow, gastrointestinal mobility and energy release) can be regarded as the basis vector. The hormones (quality dimensions) such as 'Noradrenaline', 'Dopamine' and 'Serotonin' are regarded as the projections or the degrees of freedom of the basis vectors. A emotion (concept) obtained by the collective representation of attributes from different hormones can be regarded as a quantum state obtained from its basis vectors projected along different degrees of freedom.

The aforementioned inferences convey that the high dimensional formal representation of conceptual spaces exhibits aspects of a quantum system. Following this, we model consciousness process via proposed quantum-like high dimensional formal representation of conceptual spaces to observe the quantum process in the consciousness modelling

# 4. Formal method to achieve consciousness via high dimensional formal representation of conceptual spaces:

In this section of the paper, we propose that aforementioned high dimensional formal representation of conceptual spaces achieves consciousness in a way similar to human via following procedure as shown in Fig. 3. In addition to this, we also propose an algorithm for conceptual scaling of a real world scenario to a formal context under different quality dimensions. To start, the real world scenario *R* undergoes the process of conceptual scaling that converts *R* to set of object (cues) *G* with set of attribute *M* that best describes the cue learnt at *T* instance of time granules called conceptual scales as shown in Algorithm 1. In this process of conceptual scaling, we interpret the scenario R under one perspective. We regard quality dimensions (in terms of conceptual spaces) and multiple contexts (in terms of formal representation) as perspectives. The algorithm takes a real world scenario in terms of instances that are represented by categorical propositions and introduced in advance. For example, considering apple with regard to perspective say, 'taste' would yield categorical proposition 'apple is sweet' or 'apple is sour', when considering apple with regard to another perspective say, 'colour' would yield proposition 'apple is red' or 'apple is green'. Under this illustrative scenario, the instance 'apple' will have two attributes 'sweet' and 'sour' under a perspective 'taste' while the same instance 'apple' will have two different attributes 'red' and 'green' under a different quality dimension 'colour'. As mentioned in algorithm 1, we first take a single perspective of the scenario and retrieve all associated propositions with regard to the cue in terms of a triple {cue, description, relation} at time *t*. The relation between the cue and the description holds '1' if the proposition is true with regard to the cue or '0' otherwise. For example, if the apple is sweet, then the proposition 'apple is sweet' will have 1 and 'apple is sour' will have 0. It can be observed that now the prepositions are transformed into properties with regard to the instance. Hence, each proposition conveys more information about the instance. Upon finishing the entire set of cues in first perspective, the algorithm moves to the next perspective until the completion of entire set of perspectives. This process of conceptual scale creation is called theory driven conceptual scaling (Moller, 1999).

We consider the attributes from a single quality dimension for the purpose of analysis although we have derived a quantum inspired formal representation for a high dimensional mental ensemble with multiple dimensions in the previous section. It should be noted that analysis under single quality dimension is quantum inspired as well but with respect to single particles of quantum theory as shown in Fig. 2 (Gruska & Republik, 2004)(Bruza & Busemeyer, 2012). Also, we measure the ensemble after every introduction of new cue to have a focus on the changes and understanding of knowledge caused by the cue(Aerts et al., 2013).

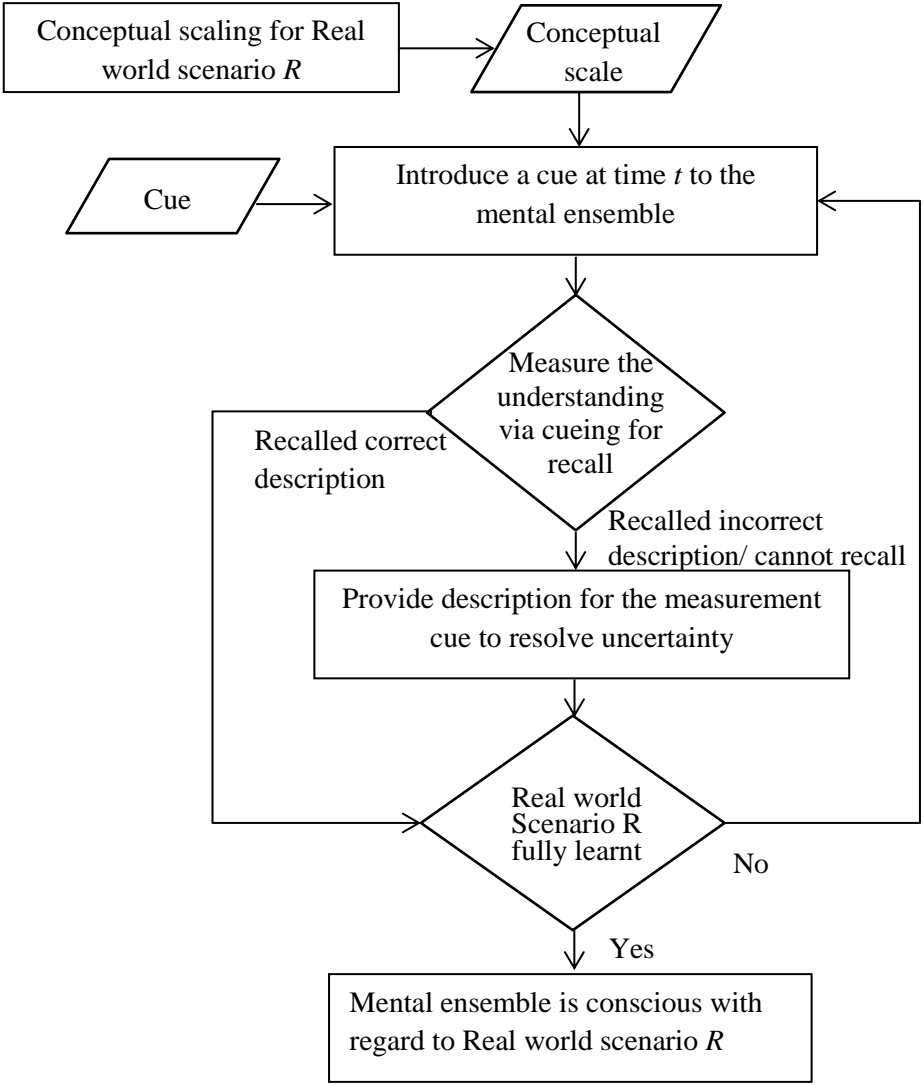

Figure 3: Consciousness via high dimensional formal representation of conceptual spaces

With the aforementioned setup, we introduce the task as cue at time granule $t_0$. Once the task is introduced to the ensemble, we measure the ensemble's understanding with a measurement cue for recalling corresponding association. This measurement cue may or may not cause an uncertainty state in the mental ensemble. We regard the state of the mental ensemble which is not able to recall the association corresponding to the measurement cue as uncertain state. If the ensemble is able to recall the association corresponding to the measurement cue, we regard this a conscious state in which the ensemble is conscious about its knowledge at the time granule $t_0$

```
Input: R := Real World Scenario
Output: K := Formal Context (cue, description, relation)
for i=1:M
    R_i:= Real world scenario described with a single quality dimension (perspective)
    W_i:= Number of categorical propositions in R_i
    for t=1: Total number of time granules
        (Instance id(t), categorical propositions on instance(t))= R_i(t)
        cue(i, index)=Instance id(t)
        do{
            description(i, j)=categorical propositions on instance(t)
            relation(i, index, j)=1 (proposition is true for the cue) or 0 otherwise
            j=j+1
        }while(j<W_i)
        Index=index+1
    end
end
```

Algorithm 1: Conceptual scaling of a real world Scenario under different quality dimensions.

If such uncertain mental state is has occurred during the process of measurement, it is resolved with presenting the association with regard to measurement cue that caused the uncertainty. The supporting cue may resolve uncertain state to definite state or may introduce further uncertainty. If there are uncertain states, it has to be resolved recursively by presenting association to the ensemble till it definite such that the ensemble is capable of recalling associations related to all the instance that were introduced till time $t_{i-1}$. In the case, there are no uncertain states in the mental ensemble; a cue can be supplied to the mental ensemble to proceed for learning. The process of providing cues with their associations has to performed recursively and correspondingly with respect to uncertain and conscious states. On providing all the intended cues if the ensemble does not have any uncertain states, the ensemble is said to attain consciousness about itself with regard to task at time $t_i$. Since the learning is subjected to consciousness, as the learning of the mental ensemble evolves, the consciousness evolves as well (Overgaard & Mogensen, 2017). The process of attaining consciousness with regard to learning task at successive time steps is shown in Fig. 3.

This aforementioned setup and procedure for a high dimensional formal representation of conceptual spaces derived in section 3 achieves consciousness by possessing knowledge about itself at a given time instant $t_i$ with method proposed in the current section in way similar to human (Van Hateren, 2018). In humans, consciousness facilitates learning further the conscious contents are stable contents. Consciousness in human can be regarded as stable understanding of a matter at a given point of time. However, consciousness evolves time to time and with a measurement or interaction with environment. Having proposed our method to achieve consciousness in conceptual spaces, we provide an experimental analysis of the same in the next section of the paper with real world scenario of learning digits based on their types to analyse their quantum aspects while attaining the consciousness.

## 5. Experimental Analysis:

In the previous section of the paper, we have proposed high dimensional formal representation of conceptual spaces that is analogous to a quantum system represented by the N qubits and a formal method to attain consciousness via proposed high dimensional conceptual space representation. In this section, we have provided the experimental analysis of the method proposed in earlier section to show that the achieved consciousness exhibits quantum aspects as well.

### 5.1 Experimental setup:

To proceed with experimental analysis, we present a learning task to the mental ensemble that follows proposed high dimensional formal representation of conceptual space. Table 2 obtained by Algorithm 1 represents the real world task of learning the types of integers presented to the mental ensemble that adapts the proposed high dimensional formal representation of conceptual spaces with temporal characteristics (Conceptual Time system). The mental ensemble learns the provided task via agent and environment interaction bases (Bruza & Busemeyer, 2012). In Table 2, it can be observe that the integers are classified under a quality dimension 'types' with five different types of 'description' of types. Although this a single quality dimension, the integer forms a conceptual space that projects the information presented to it in 5- dimensional space. The 'descriptions' are the basis vectors namely, $|u\rangle$, $|v\rangle$, $|w\rangle$, $|x\rangle$ and $|y\rangle$ representing 'Composite', 'Even', 'Odd', 'Prime' and 'Square'. Let the amplitude with regard to each respective basis states be $\alpha_0, \alpha_1, \alpha_2, \alpha_3$ and $\alpha_4$. The states are the integers namely S1, S2, S3, S4, S5, S6, S7, S8 and S9.

With this setup of the conceptual space, when an information is presented to the system as shown in the first record of Table 3, the mental ensemble develops a belief state $|\psi\rangle$ as shown in equation 12 and (a) of Fig. 5. This belief state is represented by the linear super position of basis states.

$$|\psi\rangle = \alpha_0|u\rangle + \alpha_1|v\rangle + \alpha_2|w\rangle + \alpha_3|x\rangle + \alpha_4|y\rangle \qquad (12)$$

The contribution of basis states is determined the probability amplitude of the corresponding basis state. Quantum system as well as human cognitive process develops belief states by assigning initial probabilities to the mental ensemble to proceed for learning (Bruza & Busemeyer, 2012). Since there are 5 basis states in the mental ensemble, we assign each basis state with probability amplitude of $\frac{1}{\sqrt{5}}$ so that total probability of the ensemble remains 1. Equation 12 is rewritten with assigned probability amplitudes as shown in equation 13.

$$|\psi\rangle = \frac{1}{\sqrt{5}}|u\rangle + \frac{1}{\sqrt{5}}|v\rangle + \frac{1}{\sqrt{5}}|w\rangle + \frac{1}{\sqrt{5}}|x\rangle + \frac{1}{\sqrt{5}}|y\rangle \qquad (13)$$

Table 2: Types of integers based on digits 1 to 9 for the ensemble to learn

|       | Composite | Even | Odd | Prime | Square |
|-------|-----------|------|-----|-------|--------|
| One   | 0 | 0 | 1 | 0 | 1 |
| Two   | 0 | 1 | 0 | 1 | 0 |
| Three | 0 | 0 | 1 | 1 | 0 |
| Four  | 1 | 1 | 0 | 0 | 1 |
| Five  | 0 | 0 | 1 | 1 | 0 |
| Six   | 1 | 1 | 0 | 0 | 0 |
| Seven | 0 | 0 | 1 | 1 | 0 |
| Eight | 1 | 1 | 0 | 0 | 0 |
| Nine  | 1 | 0 | 1 | 0 | 1 |

Table 3: Learning the types of integers with regard to time granule by the mental ensemble

| Time granule | Learning components | Measurement cue | Supporting cue to resolve uncertainty |
|---|---|---|---|
| 0 | Attributes are introduced to the system | Is it true that all objects have attribute composite, odd, even, prime and square? | 1 |
| 1 | Introduction of digit 1 | Is it true that all objects have attributes odd and square? | 2 |
| 2 | Introduction of digit 2 | Is it true that all object have attribute square also have odd? | 4 |
| 3 | Introduction of digit 4 | Is it true that all object have attribute prime also have even? | 3 |
| 4 | Introduction of digit 3 | Is it true that all object have attributes prime and sqaure, that also has attributes composite, even and odd | - |
| 5 | No cue | Is it true that all object has attributes even and sqaure, that it also has attribute composite? | - |
| 6 | No cue | Is it true that all object has attribute composite, that also has attributes Even, Square? | 6 |
| 7 | Introduction of digit 6 | Is it true that all object has attribute even and odd , that also has attributes composite, prime and square? | - |
| 8 | No cue | Is it true that all object has attributes composite, that also has even? | 9 |
| 9 | Introduction of digit 9 | Is it true that all object has attributes composite, odd that also has squares? | - |

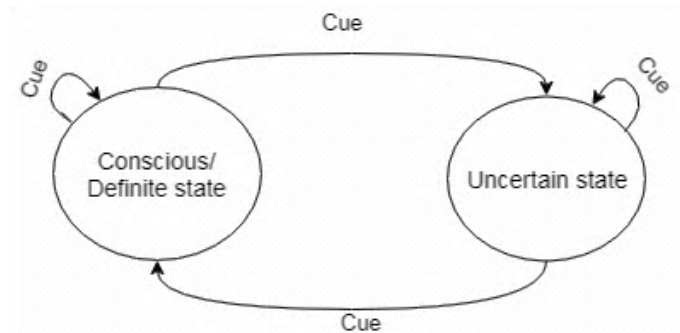

Figure 4: Life cycle of a mental state in a metal ensemble

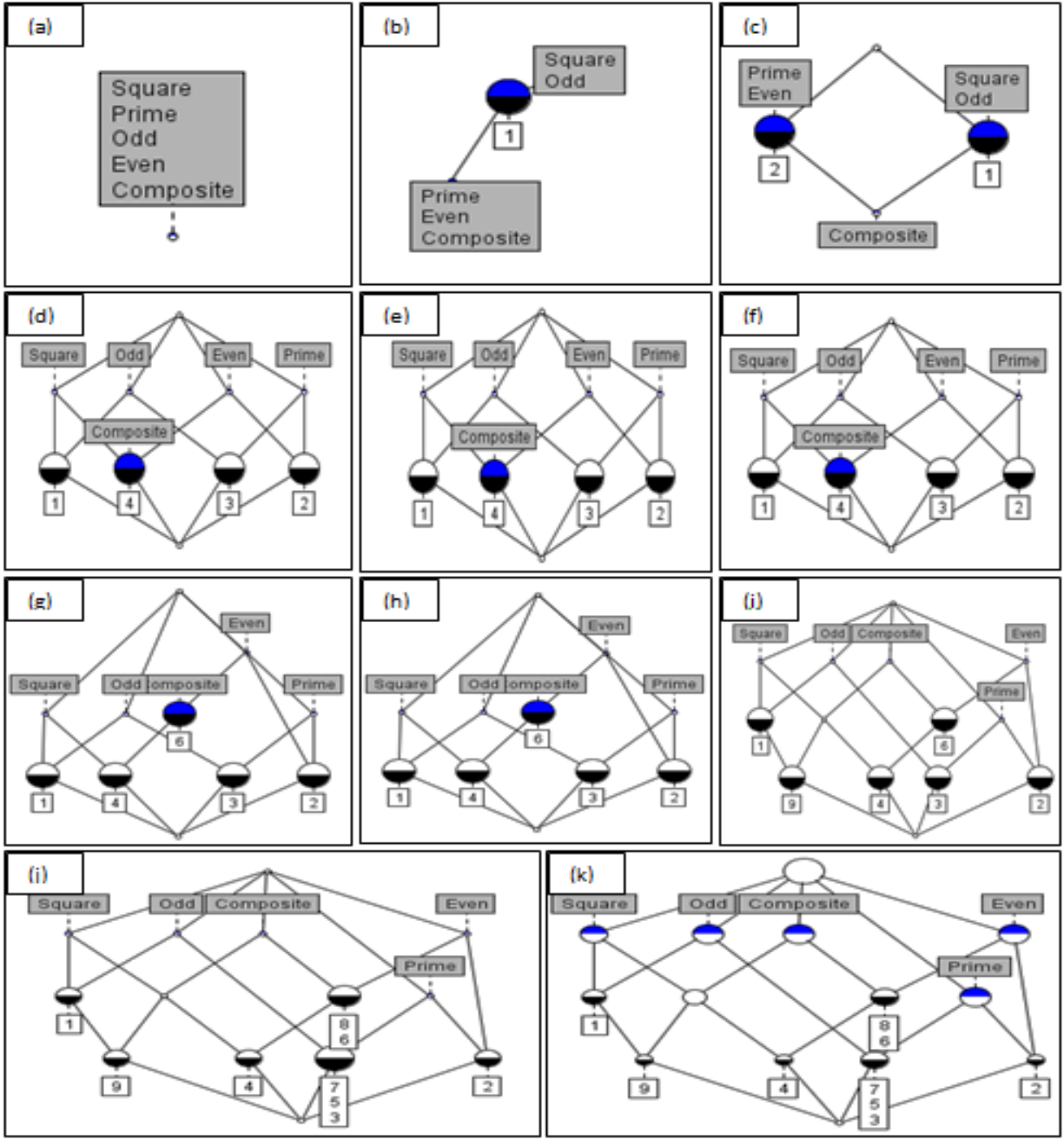

Figure 5: Changes in mental ensembles with respect to changes in mental states. Sub-figure (a)-(i) shows the changes in the mental states of the growing mental ensemble. Sub-figure(k) shows the unstable mental states.

Upon the introduction of the information at time granule 0, we measure the understanding of the mental ensemble with measurement cue. As per the current understanding of the mental ensemble, the state $|\psi\rangle$ formed a belief state. However, the ensemble is uncertain to answer if all the states that will exist in the mental ensemble will have similar probability amplitude. Since the mental ensemble learn via agent and environment interaction bases whenever there is an uncertainty, the mental ensemble makes an observation (takes a cue). In this case, integer 1 is introduces as cue to the ensemble and state $|S1\rangle$ is learnt as shown in equation 14 since $|S1\rangle$ is project along the subspaces of 'odd' and 'square' only. Similarly, the learning of states and the corresponding probability amplitude of the basis vectors is carried out as shown in Table 3. Learning is continued by learning the projection onto subspaces via updating probability amplitude with regard to basis vectors till the ensemble learns all the possible states in the learning task.

$$|S1\rangle = \frac{1}{\sqrt{10}}|w\rangle + \frac{1}{\sqrt{10}}|y\rangle \qquad (14)$$

It is interesting to note that the mental ensemble reaches conscious state (uncertain state) or remains in a conscious state (uncertain state) because of a cue or the observation from the environment as shown in Fig. 4. In order to reach a conscious state (to further recall the association of the cue) at time t, the ensemble should collapse from the superposition state. Upon learning the presented task by the mental ensemble shown in Table 2 in sequence of interaction between the ensemble and the environment, each state or integer is represented in term of projections with regard to attributes of the concerned quality dimension. Learnt mental ensemble in terms of concept lattice representing concepts in hierarchical layout is shown in Fig. 3. The (k) subfig. of Fig. 5 shows the fully learnt mental ensemble with conscious (Black and white circles) as well as redundant uncertain mental states (Blue and white circles). A conscious mental state at a given time is obtained by collapse of superposition by projection of state into subspace in terms of its probability amplitude. In Fig. 6, we have represented the conscious mental states at time granule 9 into a multi-dimensional vector space where each dimension is description about the state. In this, each projection corresponds to an attribute such as 'composite', 'odd', 'even', 'prime' and 'square' as in Table 2. If the particular digit possess the aforementioned attributes, then the concept represented the digit is projected along the attribute in the multi-dimensional space. This implies the probability amplitude with regard to basis vector (attribute of the digit) that contributes for the digit is relatively higher than the probability amplitude of the basis vector (not an attribute of the digit) that does not contribute for the digit. It can be observed from Fig. 6 that mental states '3, 5, 7' has similar description with multi-dimensional projection of states { $|S3\rangle, |S5\rangle, |S7\rangle$ } has similar projections in corresponding subspaces. Such similarity can be observed with mental states '6 and 8' and projection { $|S6\rangle, |S8\rangle$ }. It can be observed from Fig. 5, that digits with similar projections are grouped under a concept in a concept lattice.

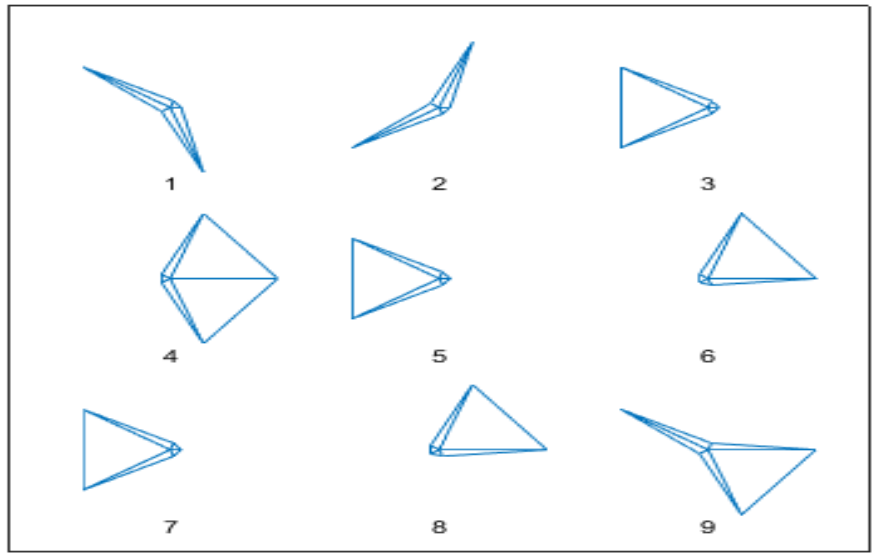

Figure 6: Projection of conscious mental states in multi-dimensional space

The model collapses from its linear superposition to respond for a cue by recalling its associations. As shown in Fig. 4, the model may become conscious or uncertain because of the cue in a way similar to quantum theories. To strengthen an association that exist between a cue and its description, the model amplifies it probability amplitude. Hence, learning in a high dimensional formal representation of conceptual spaces is learning the association between objects and attributes via learning amplitude for each projection and linear combination such projection of state on subspaces with regard to certain quality dimension in a way similar to quantum theory.

**5.2. Discussions:**

Quantum theory can be applied for macroscopic scenarios considering the cues and transitions from environment. Quantum theories are best suitable for human cognition process since human cognition cannot follow single trajectory to settle on a conclusion. Super consciousness process that happens in human brain looks for all possible trajectories to solve a problem (Mensky, n.d.). It further selects among the entire possible trajectories to solve the problem. Traditional concept theories do not provide satisfactory treatment in modelling conceptual dynamics. Quantum inspired formal representation of conceptual space with temporal characteristics is an attempt that provides better treatment in explanation conceptual dynamics in our perspective. The two fundamental principles of quantum mechanics such as superposition and entanglement is revisited with respect to quantum inspired formal representation of conceptual space. Further, measurement in quantum inspired formal representation of conceptual space plays a vital role in transitions between states in way similar to quantum theory. However, we have not addressed the compatibility issues of quantum theories. In order to study the compatibility between the mental sates in a mental ensemble, super consciousness (meta-consciousness) has to be brought in. Unlike higher order consciousness, meta-consciousness involves processing the conscious states itself for complicated cognitive processes. This may lead to several other cognitive processes such as perception and views. However, formal representation of such cognitive processes is an emerging area of research. Human cognitive subspaces are unsymmetrical in nature while the quantum subspaces are symmetrical. By adapting quantum models to cognitive process, we are left a constraint of symmetrical subspaces. In our future work, we endeavour to model compatibility issues of quantum cognition.

## 6. Conclusions:

High dimensional formal representation of conceptual spaces shows quantum aspects as per our research. Literature reports a strong association between quantum theories and cognition, in particular consciousness. In this paper, we have derived the formal representation of conceptual spaces that shows quantum aspects by vector space representation of conceptual spaces. Further, we have adapted the proposed formal representation of conceptual spaces to a constructive scenario that achieves higher order consciousness during the learning process. The temporal characteristics of the proposed model allows the mental ensemble evolve based on the interaction with the environment. Adding to these, we have proposed an algorithm for conceptual scaling of real world scenario to a formal context.

**Acknowledgement:** Authors sincerely acknowledge the support from Department of Science and Technology, Government of India under the Grant Number: SR/CSRI/118/2014.

References:

Aerts, D. (2007). Quantum Interference and Superposition in Cognition: Development of a Theory for the Disjunction of Concepts. *ArXiv*, (0705.0975), 43. Retrieved from http://arxiv.org/abs/0705.0975

Aerts, D. (2009). Quantum structure in cognition. *Journal of Mathematical Psychology*, *53*(5), 314–348. http://doi.org/10.1016/j.jmp.2009.04.005

Aerts, D., Gabora, L., & Sozzo, S. (2013). Concepts and their dynamics: A quantum-theoretic modeling of human thought. *Topics in Cognitive Science*, *5*(4), 737–772. http://doi.org/10.1111/tops.12042

Arecchi, F. T. (2011). Phenomenology of consciousness from apprehension to judgment. *Nonlinear Dynamics, Psychology, and Life Sciences*, *15*(3), 359–375. Retrieved from http://www.ncbi.nlm.nih.gov/pubmed/21645435

Arecchi, F. T. (2015). Cognition and Language: From Apprehension to Judgment—Quantum Conjectures. In *Chaos, Information Processing And Paradoxical Games: The Legacy Of John S Nicolis* (pp. 319–343). World Scientific.

Aroyo, L., & Welty, C. (2018). The Quantum Collective. In *Companion of the The Web Conference 2018 on The Web Conference 2018* (pp. 1101–1103). International World Wide Web Conferences Steering Committee.

Aswani Kumar, C., Ishwarya, M. S., & Loo, C. K. (2015). Formal concept analysis approach to cognitive functionalities of bidirectional associative memory. *Biologically Inspired Cognitive Architectures*, *12*, 20–33. http://doi.org/10.1016/j.bica.2015.04.003

Baars, B. J., & Edelman, D. B. (2012). Consciousness, biology and quantum hypotheses. *Physics of Life Reviews*, *9*(3), 285–294. http://doi.org/10.1016/j.plrev.2012.07.001

Beck, F., & Eccles, J. C. (1992). Quantum aspects of brain activity and the role of consciousness. *Proceedings of the National Academy of Sciences of the United States of America*, *89*(23), 11357–11361. http://doi.org/10.1073/pnas.89.23.11357

Belohlavek, R. (2008). Introduction to formal concept analysis. *Palacky University, Department of Computer Science, Olomouc*, 47.

Bolt, J., Coecke, B., Genovese, F., Lewis, M., Marsden, D., & Piedeleu, R. (2016). Interacting conceptual spaces. *Electronic Proceedings in Theoretical Computer Science, EPTCS*, *221*(2016), 1–30. http://doi.org/10.4204/EPTCS.221.2

Broekaert, J., & Busemeyer, J. (2018). Episodic source memory over distribution by quantum-like dynamics--a model exploration. *arXiv preprint arXiv:1806.03321*.

Bruza, P. D., & Hoenkamp, E. C. (2018). Reinforcing Trust in Autonomous Systems: A Quantum Cognitive Approach. In *Foundations of Trusted Autonomy* (pp. 215-224). Springer, Cham.

Busemeyer, J. R., & Bruza, P. D. (2012). *Quantum models of cognition and decision*. Cambridge University Press.

Busemeyer, J. R., Fakhari, P., & Kvam, P. (2017). Neural implementation of operations used in quantum cognition. *Progress in Biophysics and Molecular Biology*, *130*, 53–60. http://doi.org/10.1016/j.pbiomolbio.2017.04.007

Busemeyer, J. R. (2018). Old and new directions in strategy selection. *Journal of Behavioral Decision Making*, *31*(2), 199-202.

Busemeyer, J. R., & Wang, Z. (2014). Quantum cognition: Key issues and discussion. *Topics in Cognitive Science*, *6*(1), 43–46. http://doi.org/10.1111/tops.12074

Kitto, K. & Aliakbarzadeh, M.,(2018). Preparation and measurement in quantum memory models. *Journal of Mathematical Psychology*, *83*, 24-34.

Knill, E., Laflamme, R., Barnum, H., Dalvit, D., Dziarmaga, J., Gubernatis, J., ... & Zurek, W. H. (2002). Introduction to quantum information processing. *arXiv preprint quant-ph/0207171*.

Farrell, J., & McClelland, T. (2017). Editorial: Consciousness and Inner Awareness. *Review of Philosophy and Psychology*, *8*(1), 1–22. http://doi.org/10.1007/s13164-017-0331-x

Fields, C., Hoffman, D. D., Prakash, C., & Singh, M. (2018). Conscious agent networks: Formal analysis and application to cognition. *Cognitive Systems Research*, *47*, 186–213. http://doi.org/10.1016/j.cogsys.2017.10.003

Gärdenfors, P. (2004). *Conceptual spaces: The geometry of thought*. MIT press.

Gök, S. E., & Sayan, E. (2012). A philosophical assessment of computational models of consciousness. *Cognitive Systems Research*, *17–18*, 49–62. ttp://doi.org/10.1016/j.cogsys.2011.11.001

Gruska, J. (1999). *Quantum computing* (Vol. 2005). London: McGraw-Hill.

Gunji, Y. P., Sonoda, K., & Basios, V. (2016). Quantum cognition based on an ambiguous representation derived from a rough set approximation. *BioSystems*, *141*, 55–66. http://doi.org/10.1016/j.biosystems.2015.12.003

Hameroff, S., & Penrose, R. (2014). Consciousness in the universe: A review of the 'Orch OR' theory. *Physics of Life Reviews*, *11*(1), 39–78. http://doi.org/10.1016/J.PLREV.2013.08.002

Hameroff, S. R., Craddock, T. J. A., & Tuszynski, J. A. (2014). Quantum effects in the understanding of consciousness. *Journal of Integrative Neuroscience*, *13*(02), 229–252. http://doi.org/10.1142/S0219635214400093

Igamberdiev, A. U., & Shklovskiy-Kordi, N. E. (2017). The quantum basis of spatiotemporality in perception and consciousness. *Progress in Biophysics and Molecular Biology*, *130*, 15–25. http://doi.org/10.1016/j.pbiomolbio.2017.02.008

Kitto, K., Bruza, P. D., & Gabora, L. (2012). A Quantum Information Retrieval Approach to Memory. In *Neural Networks (IJCNN), The 2012 International Joint Conference on. IEEE*.

Kumar, C. A., Ishwarya, M. S., & Loo, C. K. (2015). *Modeling associative memories using formal concept analysis*. *Advances in Intelligent Systems and Computing* (Vol. 331). http://doi.org/10.1007/978-3-319-13153-5_11

Li, J., Zhang, P., Song, D., & Hou, Y. (2016). An adaptive contextual quantum language model. *Physica A: Statistical Mechanics and Its Applications*, *456*, 51–67. http://doi.org/10.1016/j.physa.2016.03.003

Lövheim, H. (2012). A new three-dimensional model for emotions and monoamine neurotransmitters. *Medical Hypotheses*, *78*(2), 341–348. http://doi.org/10.1016/j.mehy.2011.11.016

Masuyama, N., Loo, C. K., & Kubota, N. (2014). Quantum-Inspired Bidirectional Associative Memory for Human–Robot Communication. *International Journal of Humanoid Robotics*,

*11*(02), 1450006. http://doi.org/10.1142/S0219843614500066

Melkikh, A. V., & Khrennikov, A. (2015). Nontrivial quantum and quantum-like effects in biosystems: Unsolved questions and paradoxes. *Progress in Biophysics and Molecular Biology*, *119*(2), 137–161. http://doi.org/10.1016/j.pbiomolbio.2015.07.001

Mensky, M. B. (n.d.). Super-intuition and correlations with the future in Quantum Concept of Consciousness. *ArXiv:1407.2627v1 [Physics.Gen-Ph]*, 1–22. Retrieved from http://arxiv.org/pdf/1407.2627.pdf

Nielsen, M. A., & Chuang, I. (2002). Quantum computation and quantum information.

O'Rourke, J. (1993). Consciousness explained. *Artificial Intelligence*, *60*(2), 303–312. http://doi.org/10.1016/0004-3702(93)90006-W

Overgaard, M., & Mogensen, J. (2017). An integrative view on consciousness and introspection. *Review of Philosophy and Psychology*, *8*(1), 129–141. http://doi.org/10.1007/s13164-016-0303-6

Reggia, J. A., Katz, G., & Huang, D. W. (2016). What are the computational correlates of consciousness? *Biologically Inspired Cognitive Architectures*, *17*, 101–113. http://doi.org/10.1016/j.bica.2016.07.009

Sergioli, G., Santucci, E., Didaci, L., Miszczak, J. A., & Giuntini, R. (2018). A quantum-inspired version of the nearest mean classifier. *Soft Computing*, *22*(3), 691–705. http://doi.org/10.1007/s00500-016-2478-2

Tversky, B. (1993). COGNITIVE MAPS, COGNITIVE COLLAGES, AND SPATIAL MENTAL MODELS BARBARA. *Progress and New Trends in 3D Geoinformation Sciences Lecture Notes in Geoinformation and Cartography*, *716*(May), 1–17. http://doi.org/10.1007/3-540-57207-4

van Hateren, J. H. (2018). A theory of consciousness: computation, algorithm, and neurobiological realization. *arXiv preprint arXiv:1804.02952*.

Woff, K. E. (2011). Temporal Concept Analysis explained by examples. *CEUR Workshop Proceedings*, *757*, 104–118.

Zhang, Q., Balakrishnan, S. N., & Busemeyer, J. (2018). Fault Detection and Adaptive Parameter Estimation with Quantum Inspired Techniques and Multiple-Model Filters. In *2018 AIAA Guidance, Navigation, and Control Conference* (p. 1124).